# Experiments in Linear Template Combination using Genetic Algorithms

**Nikhilesh Bhatnagar, Radhika Mamidi**
International Institute of Information Technology, Hyderabad
nikhilesh.bhatnagar@research.iiit.ac.in, radhika.mamidi@iiit.ac.in

## ABSTRACT

Natural Language Generation systems typically have two parts - strategic ("what to say") and tactical ("how to say"). We present our experiments in building an unsupervised corpus-driven template based tactical NLG system. We consider templates as a sequence of words containing gaps. Our idea is based on the observation that templates are grammatical locally (within their textual span). We posit the construction of a sentence as a highly restricted sequence of such templates. This work is an attempt to explore the resulting search space using Genetic Algorithms to arrive at acceptable solutions. We present a baseline implementation of this approach which outputs gapped text.

## INTRODUCTION

NLG is the task of generating natural language from non-linguistic inputs. Most NLG systems can be classified into two broad camps - template based and statistical. Template based systems are generally characterized by structure gapped text (slot-filler structure) which is predominantly manually created (Reiter 1995) which generally result in high quality text but also limited linguistic coverage. Statistical systems such as (Langkilde 1998) on the other hand, use data-driven algorithms for text generation and have little-to-no reliance on repetitive manual resources making them more adaptable and maintainable, albeit with lesser text quality (Reiter 1995).

In this work, we consider generation on a sentence level and output gapped text. We define templates as gapped text which can be filled to generate textual output. A (partial) sentence is a linear sequence of such templates. Since there are a large number of choices (templates) at every step of sentence generation (sequence of templates), it naturally gives rise to a search space which contains all such sequences - grammatical and ungrammatical. The challenge is to navigate this large search space and arrive at reasonable grammatical sentences.

## RELEVANT WORK

Template based systems have been considered to have a very shallow linguistic representation and mapping of the non-linguistic data to the generated text. General consensus among the NLG community has been that template based systems are not as flexible, maintainable and expressive (linguistic coverage) as full fledged NLG systems (Reiter 1995).

(Reiter 1997) mentions that template based systems and NLG systems are "turing equivalent" meaning that at least in terms of expressiveness, there is no theoretical disparity between the two. Early template based systems (Kukich 1983) used large phrasal units as templates to generate text. However, later systems (van Deemter and Odijk, 1997, Theune, 2001) in addition to working with smaller templates also exhibit templates with gaps (slots) which can be filled recursively which made the systems a lot more maintainable and robust to output text requirement changes. NaturalOWL (Androustopoulos, 2007) is a template based system which generates descriptions of artifacts in a museum based on the user expertise using an OWL ontology as its data source. (Deemter 2005) gives convincing arguments that the line between template-based systems and NLG is quite blurry. (Kondadadi, 2013) present a hybrid NLG system which generates text by ranking tagged clusters of templates. Combining that with the huge amount of text data available today, corpus based template approaches seem feasible.

In this work, we focus on template structure and combination. A template is gapped text extracted from a corpus. Eg. "in the NN", "The JJ NN", etc. We observe that a template is grammatical locally, within its span. Thus, it doesn't need to be "generated" itself. It follows that linear template combination (juxtaposition) can then be used to generate sentences. The problem is then two fold: how to determine what templates to combine and constrain that with a measure of grammaticality of the generated partial sentence. We make use of NGram models to address both problems. Because the templates are grammatical locally, we can use the NGram probabilities at the template edges (junction) to discriminate whether two templates can be combined or not. Also, we use an abstraction of the template - a "syntactic signature" to compute the grammaticality of the partial sentence.

Since the search space for the partial sentences (template sequences) contains all the permutations of all lengths, it is also called a permutation space. Since we do not have a reliable "best" sentence metric, we use GAs to explore the search space to get at a "good enough" solution. We describe how the template combination and sentence grammaticality are used in the genetic algorithm in the following sections.

**APPROACH**

**Template structure and extraction**

In this paper, we restrict ourselves to linear, sub-sentential templates. To reduce their number, we consider chunks as templates. It also helps that chunks are a linguistically contained structure. We chunk the UkWaC corpus (Ferraresi 2008) (first 10M sentences) using CRFChunker (Xuan-Hieu Phan, 2006) for extracting the templates.

**Template Factoring**

In this step, we introduce gaps in the templates. Instead of the gaps being blanks, we specify some linguistic features for the gaps so that not all information is abstracted away. These features are called factors. For the baseline, we use the part-of-speech as the factor. We use the following strategies for factoring -
- Absolute (count threshold): All tokens(words in a sequence) with counts below a threshold are replaced with their factor (POS).
- Relative (rank threshold): All tokens(words in a sequence) with rank below a threshold are replaced with their factor (POS).

Obviously, more aggressive the factoring, more abstract are the templates.

# GENETIC ALGORITHM

Genetic algorithms are a class of optimization algorithms which mimic a natural process namely natural selection (GA primer ref). Following is a general working of a GA:
1. Initialize starting population.
2. While stopping criterion is false, do 3 - 6
3. Selection: Select a sample from the population for reproduction.
4. Crossover: Derive the offspring from the parent sample.
5. Mutation: Mutate the offsprings given the mutation probability.
6. Selection: Select fit individuals from the parent and offspring populations for the next generation.

These components are described below.

## Chromosome

Each partial solution in the space is an ordered list of templates. That is the chromosome. The collection of chromosomes is the total population. We used four factoring strategies: count < 100, count < 1000, rank > 1000 and rank > 100. We extracted 18M unique templates from unfactored text and 15M, 12M, 4.6M and 1M unique templates for factored text respectively.

## Crossover

The template combination problem is addressed in the crossover function. Crossover determines how and with what probability two parent chromosomes produce an offspring. We use juxtaposition as the crossover process. So, if two templates "in the NN" and "VBZ a NN" are combined, the resulting offspring is "in the NN VBZ a NN". To determine the crossover probability, we train a trigram model on the corresponding factored text with modified Kneser-Ney smoothing (Sundermeyer 2011) using SRILM (Stolcke 2002). The crossover probability is the trigram probability of the token/factors at the junction of the two partial sentences.

## Mutation

To mutate a chromosome, we replace it with a single template chosen randomly from the total population. Note that the chromosome being replaced can have multiple templates in it. That helps curb the sentence length increase rate. The mutation probability is 0.05.

## Fitness function

The sentence grammaticality issue is handled in the fitness function. The fitness function determines how "good" a particular solution is. It is critical for selecting "good" candidates each generation, which helps converge to a set of "good" solutions.

There are a few issues for the fitness function to address -
- Variable length - grammaticality does not depend on length
- Partial sentences - extra incentive to fully formed sentences

Since the templates are locally grammatical, we simplify the sentential grammaticality to a sequence of template-level features - the syntactic signature. For the baseline, use use the chunk tags as the syntactic signature. We train a 5-gram model over the sequence of chunk tags using modified Kneser Ney smoorthing. The fitness value of a partial sentence is the total probability of the chunk tag sequence of the templates constituting it. The crossover probability is the total probability of the chunk tag sequence of the templates constituting it normalized by the length of the chromosome and divided by difference of the target sentence length.

$$P_{crossover} = 10 \char`\^ \max(P_{partial} / length_{partial}, P_{total} / (length_{partial} + 2) / abs(L - length_{partial})$$

$length_{partial}$ is the length of the chromosome
L is the length of the solution chromosomes.
Total probability is computed by considering begin and end of sentence tags as well.
We take the max so that we can distinguish between partial or a full (total) sentence.

**Evolution policy**

We use a basic tournament policy with n-best selection. To do that, a tournament is conducted between a small sample of the population (tournament size say, 10) and the n-best (fittest) offspring are selected. This is done until enough offspring are created. We use 10 as the tournament size and select the 10 best offspring from each tournament. We maintained 1M as the population size with a 50-50 parent-offspring split and ran the search for 100 generations.

**RESULTS**

Unfactored:
 1. the muslim salutation is so well drawn unanticipated changes hoping to wring community pool timebank a bbc rob & bev

Rank - 100:
 1. the NN UH 1 EX EX

Rank - 1000:
 1. two RB related NNS tell the NNP NNS 1 john 5 , 20 NNP the same time NNS � CD and plus � 1 ' NNS and NNS , telephone , NN young NNP school NNS

Count - 1000:
 1. the granary holiday cottages NNP helps to sit a JJ cray NNP supercomputer internationally of these other gentlemen com NNP magazine

Count - 100
 1. sometimes western countries may also have been located her the personal organisation skills had not been tested or used a window dialogue box should be properly studied such a safe labour seat

**OBSERVATIONS**

We observe that the fitness values fluctuate around 0.18 for all factoring experiments (that is because the syntactic signature remains the same). The chromosome length did stabilize around generation 40. Clearly, the sentences don't end with punctuations. Long distance dependencies are not handled well by NGram models, so that may not be a good grammaticality measure.

**CONCLUSIONS AND FUTURE WORK**

Clearly, we can observe that the fitness function needs improement to better reflect grammaticality. For future work, we see multiple areas of improvement:
1. Length desensitization to N in the fitness function - percentile measures We currently divide logprob by length.
2. Check fitness function behavior with correct sentences to help tune parameters.
3. Make selection depend on average length of the sentences.
4. Human evaluation and BLEU scores can be incorporated into fitness function as well.
5. Create better syntactic signatures and use a factored language model, perhaps.
6. Use semantic categorization factors in crossover probability to make the selection more robust.
7. Explore using skip-grams for grammaticality.
8. Explore methods to distinguish partial sentences from full sentences in the fitness function, and use that as a stopping criterion in tournament selection.